\documentclass{article}

\usepackage{PRIMEarxiv}

\usepackage[utf8]{inputenc} 
\usepackage[T1]{fontenc}    
\usepackage{hyperref}       
\usepackage{url}            
\usepackage{booktabs}       
\usepackage{amsfonts}       
\usepackage{nicefrac}       
\usepackage{microtype}      
\usepackage{lipsum}
\usepackage{fancyhdr}       
\usepackage{graphicx}       
\graphicspath{{media/}}     
\usepackage{gensymb}
\usepackage{enumerate}
\usepackage{multirow}
\usepackage{amsthm}
\usepackage{subfig}

\title{Sorting of Smartphone Components for Recycling Through Convolutional Neural Networks
}

\author{
  Álvaro G. Becker$^1$, Marcelo P. Cenci$^2$, Thiago L. T. da Silveira$^1$, Hugo M. Veit$^2$ \\
  $^1$ Institute of Informatics, Federal University of Rio Grande do Sul, Brazil \\
  $^2$ LACOR -- Department of Materials, Federal University of Rio Grande do Sul,  Brazil \\
  \texttt{\{agbecker,tltsilveira\}@inf.ufrgs.br, \{marcelo.cenci,hugo.veit\}@ufrgs.br} \\
}

\begin{document}
\maketitle

\begin{abstract}
 The recycling of waste electrical and electronic equipment is an essential tool in allowing for a circular economy, presenting the potential for significant environmental and economic gain. However, traditional material separation techniques, based on physical and chemical processes, require substantial investment and do not apply to all cases. In this work, we investigate using an image classification neural network as a potential means to control an automated material separation process in treating smartphone waste, acting as a more efficient, less costly, and more widely applicable alternative to existing tools. We produced a dataset with 1,127 images of pyrolyzed smartphone components, which was then used to train and assess a VGG-16 image classification model. The model achieved 83.33\% accuracy, lending credence to the viability of using such a neural network in material separation.
 \end{abstract}


\section{Introduction}
In a report released by the United Nations University (UNU) in 2020, the global generation of waste electrical and electronic equipment (WEEE) was estimated at 53.6 million tons annually, or 7.3 kg per capita, with WEEE being the fastest-growing solid waste stream in recent years (from 9.2 million tons in 2014 to a projected 74.7 million tons annually by 2030)~\cite{e-waste2020}. 
The context of WEEE generation also includes a high degree of informality in end-of-life management, with only 17.4\% being properly documented and disposed of through formal means, primarily due to technological challenges in collection and recycling faced by the actors involved in this process~\cite{e-waste2020}. 
From this scenario, the report emphasizes that recycling is a fundamental strategy for minimizing the environmental and societal impacts of the WEEE generation, as it is an essential component of the 2030 Agenda for Sustainable Development under the following United Nations Sustainable Development Goals: Goal 3 (Good Health and Well-being), Goal 6 (Clean Water and Sanitation), Goal 8 (Decent Work and Economic Growth), Goal 11 (Sustainable Cities and Communities), Goal 12 (Responsible Consumption and Production), and Goal 14 (Life Below Water).

Over the past decade, there has been a concentration of scientific efforts to find recycling solutions for WEEE. Typically, methods established in the metallurgical industry are adapted for WEEE processing. It is the case of the company Umicore, considered a global benchmark in the field, which has its processes based on copper and lead metallurgy, adding only 15\% of WEEE to the primary ores and recovering only the most precious metals, such as gold and silver~\cite{umicoree-scrap,industrialscale}. At the base of the recycling chain, the collection and handling of WEEE are still carried out using inefficient methods, with a predominance of manual labor and massive wastage of valuable components~\cite{whatdrives,wasteinbrazil}.

\begin{figure}[!t]
\centering
\includegraphics[width=2.5in]{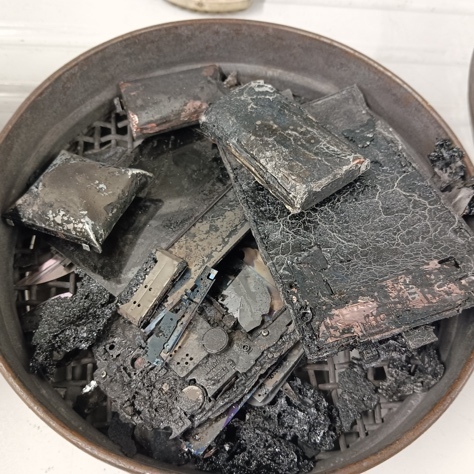}
\caption{Pyrolyzed smartphones for battery separation.}
\label{fig_pyrolysis}
\end{figure}

The ongoing work presented here stems from research efforts to increase the efficiency of processes and valorization of WEEE (in this case, applied to smartphones) in the early stages of recycling, where waste is managed by small and medium-sized recyclers lacking appropriate technology for these tasks. One of the main objectives is to minimize the need for manual handling of WEEE by automating the processes. By improving the efficiency of the early beneficiation stages (collection, sorting, and pretreatments), the downstream metallurgy processes are expected to be positively impacted in terms of material recovery.

The project begins with processing whole smartphones in pyrolysis furnaces (degradation of polymers in the absence of oxygen, generating byproducts with high energy value) to open the devices and release the components that will be subsequently separated. This approach considers all the components in the recycling chain, and no material is wasted. In Fig.~\ref{fig_pyrolysis}, degraded smartphones resulting from pyrolysis are shown. Following the pyrolysis process, a critical step is separating batteries from other electronic components (screens, printed circuit boards (PCBs), and metal parts) because the methods for recovering materials from batteries differ from those used for other components. Batteries of smartphones contain high concentrations of lithium, which is considered a critical and strategic material for many nations and companies due to the limited availability of primary ores and limited international supply~\cite{ecofriendly}. The recycling of batteries emerged as a strategy to mitigate this problem, but specific methods of recycling are needed. If the batteries are not separated from the other WEEE, the lithium is lost in the slag of the typical pyrometallurgical processes, or diluted among other elements, hindering hydrometallurgical approaches~\cite{recoverymetals}~\cite{pyro}.

At this point, the challenge of separating batteries was addressed through a machine learning strategy for image-based component separation, which has the potential to achieve efficient separation without hindering the subsequent recycling processes through metallurgical separation processes. The general idea is to allow the industry to perform these activities automatically, in which a detector and a mechanical sorting device could be coupled to a conveyor belt to separate components.


The rest of this paper is organized as follows. Section \ref{sec_related} reviews the few related works that aim to sort waste using automated learning-based solutions. The proposed methodology is described in Section~\ref{sec_method}. The results are presented and discussed in Section~\ref{sec_results}. Section~\ref{sec_conclusions} concludes this paper and indicates future works.


\section{Related Works} \label{sec_related}

Image-based waste separation has gained traction recently~\cite{computervisionwaste,classificationsystem,transferlearning,realtime,automaticsorting}, with predominant applications in separating urban waste such as paper, plastic, glass, and metals.

Lu and Chen~\cite{computervisionwaste}, in a literature review from 2022, found eight studies applying artificial neural networks for waste separation. 
Traditional backbones are trained and applied to the waste separation task in most works found.
Bobulski and Kubanek~\cite{classificationsystem} developed a convolutional neural network (CNN) to segregate different types of plastics for recycling, achieving an accuracy of over 99\%. Zhang et al.~\cite{transferlearning} used a DenseNet169~\cite{densenet} to segregate different kinds of household waste (glass, paper, textiles, metals, and plastics) and achieved 82\% accuracy in their tests. 

For WEEE applications, some studies specific to the field were published between 2022 and 2023. Yang et al.~\cite{realtime} applied the YOLOv4 network~\cite{yolov4} to classify WEEE that potentially have internal batteries from those that do not, such as laptops and printers, achieving an accuracy of 90.1\%. Lu et al.~\cite{automaticsorting} used the YOLOv3 network~\cite{yolov3} to detect and separate previously disconnected electronic components from PCBs, such as capacitors and transistors, with accuracies exceeding 90\%.

To the best of our knowledge, the present work is the first to use CNNs to classify WEEE dismantled components focusing on recycling.
We note that sorting pyrolyzed components turns out to be a challenging task since the individual components lack characteristic texture and shape. 
Separating these components (screens, PCBs, metals, and batteries) is a critical step in the recycling routes for most WEEE, and automating this step, besides being an operational and academic innovation, can generate significant increases in profitability and material recovery efficiency. 

Specifically, we aim to achieve high accuracy in battery separation, creating a material stream concentrated in batteries. In other words, to prevent other components from being misclassified as batteries and to prevent batteries from being misclassified as other components.

\section{Methodology} \label{sec_method}

\begin{figure}[!t]
\centering
\includegraphics[width=2.2in]{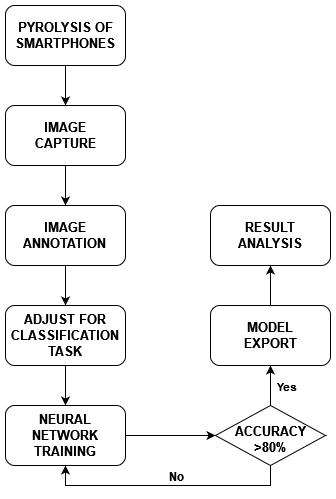}
\caption{Work steps. The pipeline includes material preparation, dataset construction, and classification model training and evaluation.}
\label{fig:flowchart}
\end{figure}

To carry out the project associated with this paper, 123 smartphones were gathered through collecting campaigns and the support of partner companies and research projects. In addition, 27 detached smartphone batteries were also gathered with support of partners. No restrictions were imposed on any characteristics of the devices, except for the requirement that they adhere to the smartphone-style. The following sections provide a detailed overview of the methodological steps applied here, which can be observed in Fig.~\ref{fig:flowchart}.


\subsection{Smartphone pyrolysis}

The smartphones were processed in a batch electric resistance furnace. The process conditions were as follows: nitrogen atmosphere, temperature of 600°C, ambient pressure, heating rate of 300°C/h, and residence time of 1 hour. Fig.~\ref{fig_pyrolysis} provides examples of the materials that result from the process. The material was also submitted to a screening with a 2cm opening to remove small particles. These small particles, after the battery separation, can be sent back to the flow of other components (PCBs, glass, and screens).

The technique of pyrolysis, applied to WEEE, generates various benefits to the downstream recycling chain. By the degradation of polymers, encapsulated metals are liberated, and the mass concentration of valuable materials increases, facilitating the posterior chemical attacks. In addition, magnets are demagnetized, the total mass to be processed decreases, metals are kept in their reduced form (preventing oxidation), and energy valuable liquids and gases are generated. The technique of pyrolysis is a hotspot in recent research, being applied to many WEEE as the first step of recycling routes. Pyrolysis furnaces are commonly used in the treatment of wastes, especially for organic wastes, and are expected to conquer more space in the WEEE recycling industry in the middle term.


\subsection{Image capture}

The image capture took place with the material randomly selected. The decision was made to position various components in a single image to facilitate future study of the image-based detection task. The present article, however, tackles the multiclass classification problem.

\begin{figure}[!t]
\centering
\includegraphics[width=2.8in]{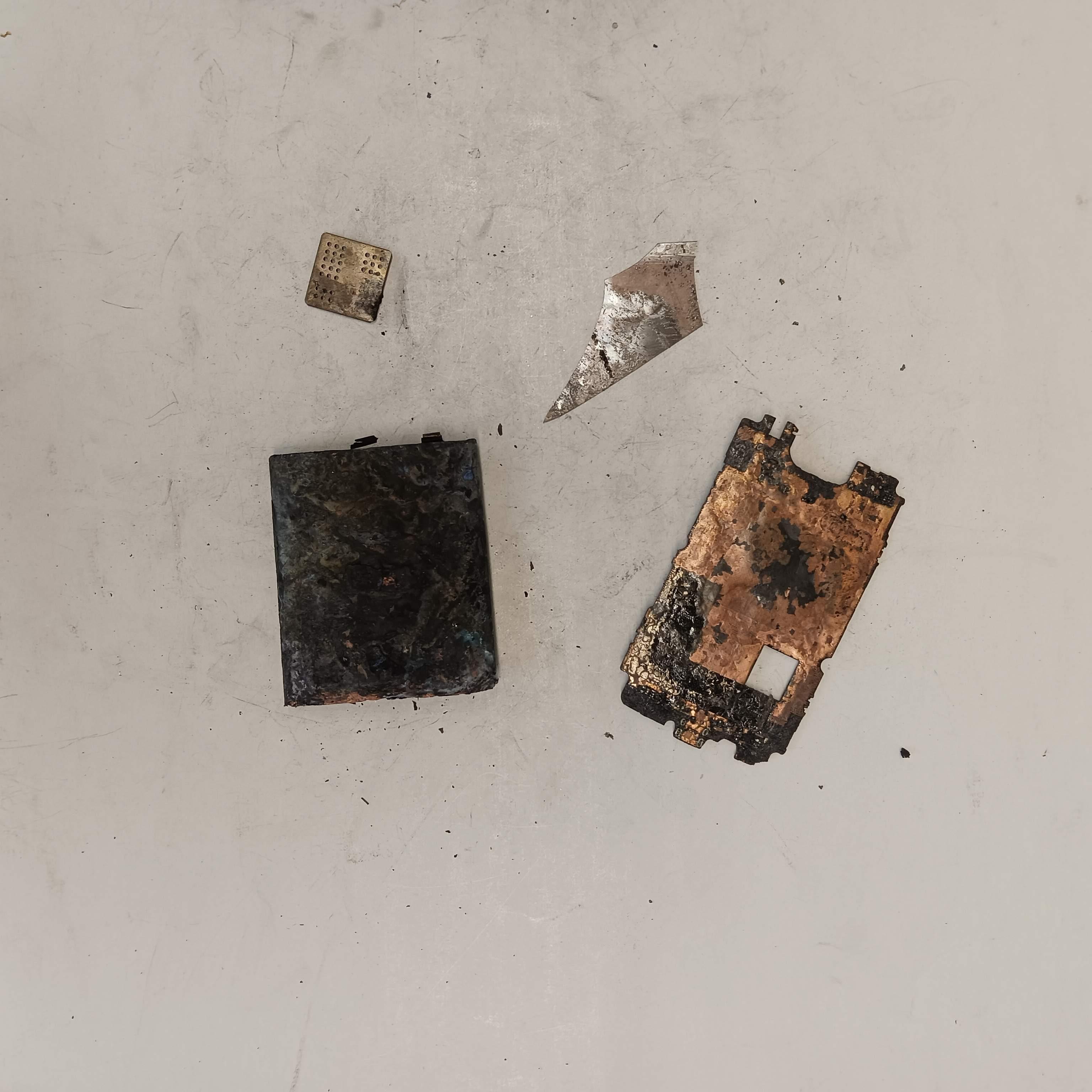}
\caption{Sample image of components on a white background. The components are samples of glass, PCB, battery, and metal piece, from the top right on, clockwise.}
\label{fig:components}
\end{figure}

In total, 300 $3072 \times 3072$ colored images were captured. Each image contains at least one battery, one piece of metal or PCB, and another random piece. The pieces were randomly placed on a background prepared in gray, black, or white (in equal proportions) under variable ambient lighting (no artificial illumination is used). The images are captured perpendicularly, roughly 50 cm from the background. Fig.~\ref{fig:components} shows an example image captured on a white background with one battery, one PCB, a small piece of glass, and another small metal piece.

After capturing the images, the components were flipped to display the reverse side, their positions were swapped, and another shot was captured. Thus, for each component, two images are captured, one for each face, without duplication. After capturing both images, the components were stored separately to prevent repetition.

\subsection{Image annotation and pre-processing}

The images were annotated using the Roboflow platform~\cite{roboflow}. All components in each image were manually outlined with a polygonal drawing tool. The components were then assigned one of four classes: Metal Piece, Battery, Glass, or Printed Circuit Board (PCB). These four classes are omnipresent in every smartphone. Indeed, after pyrolysis, they are visually the only ones to remain in gross granulometry along with the batteries.


Very small components in some pictures were not annotated, as they were considered to have little or negative contribution to learning. They were instead left as part of the background, acting as visual noise elements that will also be present in real-life applications.

\begin{figure*}[!t]
\centering
\subfloat[Metal Piece]{\includegraphics[width=1.5in]{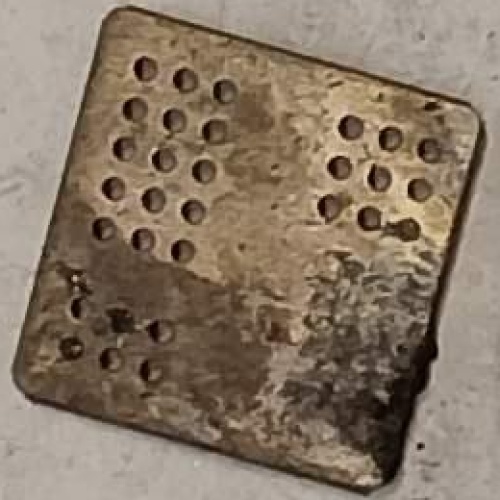}%
\label{ex_al}}
\hfil
\subfloat[Battery]{\includegraphics[width=1.5in]{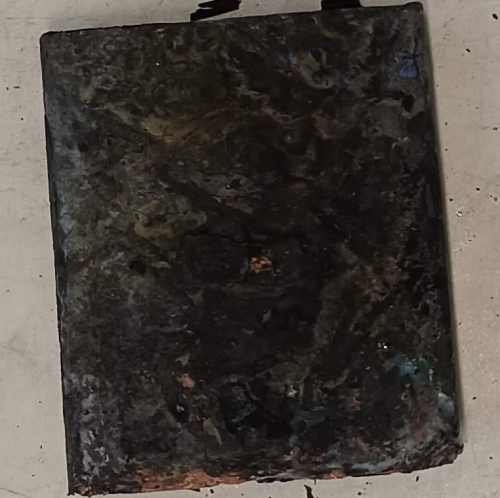}%
\label{ex_bat}}
\hfil
\subfloat[PCB]{\includegraphics[width=1.5in]{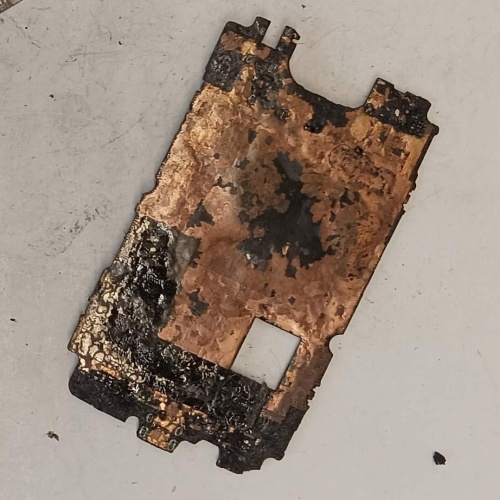}%
\label{ex_pcb}}
\hfil
\subfloat[Glass]{\includegraphics[width=1.5in]{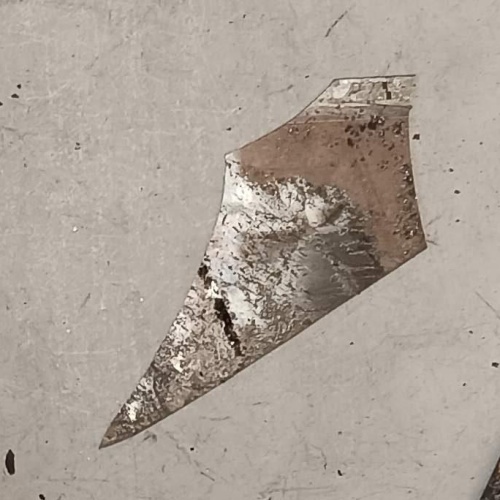}%
\label{ex_glass}}
\caption{Individual component images extracted from Fig. \ref{fig:components}.}
\label{fig:cropped_images}
\end{figure*}

The annotated images were exported in an oriented bounding box (OBB) format~\cite{yi2021oriented}. Then, for each component, a square, horizontal bounding box was circumscribed about the component's original OBB. The contents of these square BBs were then exported and resized to $500\times500$ pixels, thus composing a dataset of individual component images. Fig.~\ref{fig:cropped_images} illustrates this process for the components shown in Fig.~\ref{fig:components}.

In total, the dataset contained 1,127 images. The images were split in a 70:20:10 ratio between training, validation, and test sets, respectively, with the split happening after the cropping process, given the uneven distribution of different components within the original images. The distribution of these images among the different classes and subsets is shown in Table~\ref{tabela_fotos}.
Note that, due to the small size of the dataset, we avoided discarding any component images, despite the uneven quantities of different materials present in the images. Because of this, the entire dataset is slightly unbalanced, presenting 19.25\%, 21.83\%, 26.62\%, and  32.30\% of metal piece, PCB, battery, and glass instances, respectively.

\begin{table}[t]
\renewcommand{\arraystretch}{1.3}
    \caption{Dataset Image Distribution}
    \label{tabela_fotos}
    \begin{center}
    \begin{tabular}{c | c c c c|c}
        \hline
        Set & Metal Piece & Battery & PCB & Glass & Total\\ 
        \hline
        Training & 154 & 210 & 174 & 256 & 794\\ 
        Validation & 42 & 60 & 48 & 72 & 222\\
        Test  & 21 & 30 & 24 & 36 & 111\\ 
            \hline
        Total & 217 & 300 & 246 & 364 & 1127\\ 
        \hline
    \end{tabular}
    \end{center}

\end{table}

\par

\subsection{Neural network training}

For the desired image classification model, the VGG-16~\cite{vgg} architecture was used as a backbone, being chosen due to its shallow depth being appropriate for the small dataset used~\cite{smalldataset}. We only adjust the last VGG-16 layer to the desired number of classes.

The network we used was pre-trained on the ImageNet dataset~\cite{imagenet}. Our ablation study conducted the impact of pre-training, which significantly improves results, as discussed later.
To account for the small dataset, data augmentation was used.
A random combination of transformations was applied to each image for each training epoch.
More precisely, we considered:
\begin{enumerate}[(i)]
    \item Rotation, within $\pm 45 \degree$;
    \item Shear, within $\pm 5 \degree$;
    \item Zoom, up to 20\%;
    \item Channel shifts, within $\pm 10$;
    \item Horizontal flips;
    \item Vertical flips.
\end{enumerate}

The shear, zoom, and channel shift transformations were constrained within narrow value ranges 
so as to not deform the images to the point of impairing learning.

The training was conducted on an NVIDIA 4070 Ti GPU. The training was set to run for a maximum of 100 epochs, with early stopping enabled with a patience value of 10 epochs. This condition monitored the validation accuracy so as to export the model from the epoch where it achieved its highest value in that time frame. Batch size 32 was used, with categorical cross entropy as the loss function and the Adam optimizer with a learning rate $10^{-3}$.

\section{Results and Disscussion} \label{sec_results}

%
%

\subsection{Training results}
\label{trainingresults}

In this paper, we consider the overall accuracy,  and per-class precision and recall as figures of merit. The training accuracy is calculated as
\begin{equation}
    A = \frac{N_{True}}{N_{Total}},
\end{equation}
where $N_{True}$ is the number of correct predictions by the model and $N_{Total}$ is the total number of predictions made by the model.

The precision of a given class is calculated as
\begin{equation}
    P = \frac{N_{TP}}{N_{TP} + N_{FP}}
\end{equation}
and recall as
\begin{equation}
    R = \frac{N_{TP}}{N_{TP} + N_{FN}},
\end{equation}
where $N_{TP}$ is the number of true positives guessed by the model (that is, elements of the class in question which were correctly labeled as such), $N_{FP}$ is the number of false positives (elements incorrectly labeled as belonging to that class), and $N_{FN}$ is the number of false negatives (elements belonging to that class which were labeled as something else).

\begin{figure}[!t]
\centering
\includegraphics[width=3.2in]{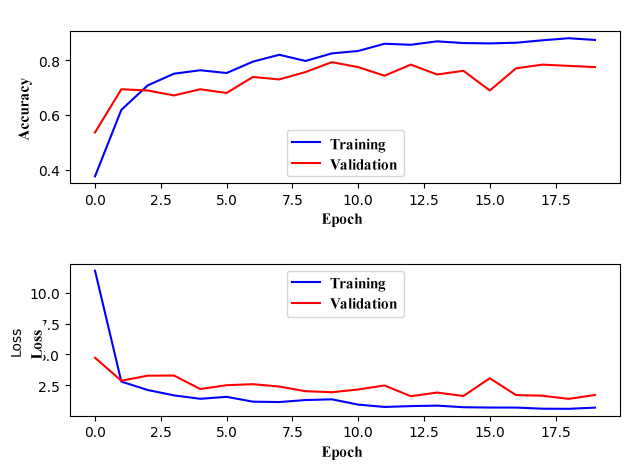}
\caption{Accuracy graphs (upper) and losses (bottom) of training and validation along the training.}
\label{fig:grafTreino}
\end{figure}

The resulting model was trained for 20 epochs (due to the early stopping), with epoch 10 yielding the best results. It achieved a training accuracy of 82.49\% and a loss of 1.3541. Meanwhile, the validation accuracy reached 79.28\%, with a validation loss of 1.9309. Fig.~\ref{fig:grafTreino} shows the graphs for Accuracy and Loss over time for training and accuracy metrics.

Running an inference test with the model on the test dataset resulted in a mean precision of 77.29\% and a mean recall of 77.35\%. The confusion matrix of this test is given in Table \ref{matriz_confusao}, while the precision and recall metrics for each class are shown in Table \ref{precision_recall}.

These results show that the network did gain the ability to generalize what it learned, and show promise for even more robust results from this classification network if trained on a larger dataset, given the small size of the one used. It should also be noted that the high contrast in precision and recall values between classes is likely attributable to the unbalanced nature of the dataset, seeing as the lowest values are associated with the two least represented classes (Metal Piece and PCB). Expanding the dataset in subsequent works would allow for eventual discarding of images so as to test training the model with a balanced dataset.

Regarding the materials flow for recycling, the model's performance was surprisingly successful, even more so considering that many opportunities for improvements and alternative tests are still to be considered. It is essential to highlight that the study's primary goal is to separate batteries. Thus, the results of precision (90.32\%) and recall (93.33\%) for the class of batteries determine the study's success.

Of the thirty tested batteries, twenty-eight were correctly identified as batteries, and two were considered PCBs (93.33\% of recall). As stated, batteries must be separated to recover lithium through specific metallurgical processes~\cite{pyro}. Batteries falling into the flow of PCBs preclude the recovery of this element. In the flow of the class of batteries, there were 28 batteries, two metal pieces, and one PCB (precision of 90.32\%). The two metal pieces in the flow of batteries do not hinder the recyclability of any of the two materials. Indeed, batteries contain a metal casing~\cite{metalcasing}, easily treated in the typical recycling route of batteries. The one PCB in the flow of batteries means that the valuables materials present at the PCBs (gold, silver, and others) would not be recovered in their typical recycling route. Finally, the model efficiently removed the glass from other flows of materials (83.33\% of recall). This is an important outcome, considering that glass is not a valuable material and only increases the mass of the materials needing treatment~\cite{glass}.

\begin{table}[t]
    \centering
    \caption{Confusion matrix of the model's test predictions}
    \label{matriz_confusao}
    \begin{tabular}{c|ccccc}
    \cline{3-6}
    \multicolumn{1}{c}{} &\multicolumn{1}{c}{} &\multicolumn{4}{c}{Prediction} \\ 
    \cline{3-6}
    \multicolumn{1}{c}{} & 
    \multicolumn{1}{c}{} & 
    \multicolumn{1}{c}{Metal Piece} & 
    \multicolumn{1}{c}{Battery} &
    \multicolumn{1}{c}{PCB} &
    \multicolumn{1}{c}{Glass} \\ \hline
    \multirow{4}{*}{\rotatebox{90}{Actual\,}}
    & \multicolumn{1}{c|}{Metal Piece}  & 13 & 2 & 6 & 0
    \\ 
    & \multicolumn{1}{c|}{Battery}  &0   & 28 &2 &0 \\   & \multicolumn{1}{c|}{PCB}  &5   & 1 & 17 &1 \\     & \multicolumn{1}{c|}{Glass}  &4   & 0 & 2 &30 \\ \hline
    \end{tabular}
    \label{tab:matconf16}
\end{table}

\begin{table}[t]
\renewcommand{\arraystretch}{1.3}
    \caption{Precision and recall for inference in each class}
    \label{precision_recall}
    \begin{center}
    \begin{tabular}{c|c|c}
        \hline
        Class & Precision & Recall\\
        \hline
        Metal Piece & 59.09\% & 61.90\%\\ 
        Battery & 90.32\% & 93.33\% \\
        PCB  & 62.96\% & 70.83\%\\ 
        Glass & 96.77\% & 83.33\%\\
        \hline
    \end{tabular}
    \end{center}

\end{table}

\subsection{Ablation studies}
\subsubsection{Pre-training}

Because the background of the images in our dataset is quite simple (nearly smooth white, gray, or black surface) and the pyrolyzed materials do not present color variation, we questioned the necessity of pre-training. It is reasonably assumed that pre-training the model on large datasets such as ImageNet~\cite{imagenet} might be excessive for this setup problem.

As such, to verify the validity of using a pre-trained network, a version of the model was trained with randomly initialized weights to compare its performance with the pre-trained model. All the training conditions were identical between models other than the initial weights.

\begin{figure}[!t]
\centering
\includegraphics[width=3.2in]{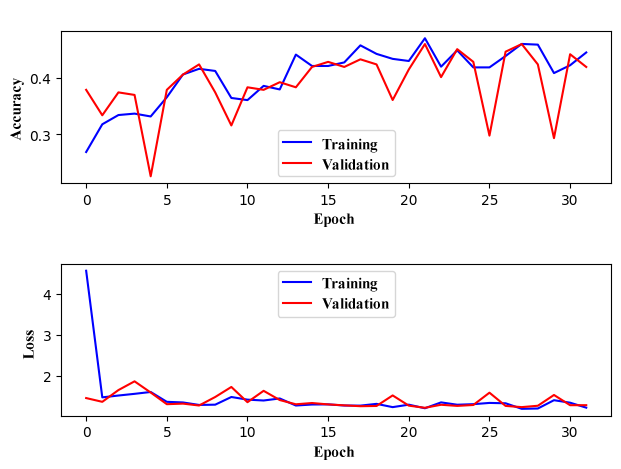}
\caption{Accuracy graphs (upper) and losses (bottom) of training and validation along the training.}
\label{fig:vgg_sem_pre_treino}
\end{figure}

The randomly initialized model trained for 32 epochs, achieving its best results at epoch 22, with a training accuracy of 46.98\% and loss of 1.2000, and a validation accuracy of 45.95\% and loss of 1.2115. 
The training graphs are shown in Fig.~\ref{fig:vgg_sem_pre_treino}, with the accuracy graph showing its learning process was inconsistent.

\par

In the inference evaluation with the test dataset, the model had a mean precision of 38.71\% and recall of 40.49\%. In fact, it did not label a single metal piece component correctly, as can be seen in Table~\ref{matriz_confusao_fromscratch}. The precision and recall for each class is shown in Table~\ref{precision_recall_fromscratch}.

\begin{table}[t]
    \centering
    \caption{Confusion matrix of the randomly initialized model's test predictions}
    \label{matriz_confusao_fromscratch}
    \begin{tabular}{c|ccccc}
    \cline{3-6}
    \multicolumn{1}{c}{} &\multicolumn{1}{c}{} &\multicolumn{4}{c}{Prediction} \\ 
    \cline{3-6}
    \multicolumn{1}{c}{} & 
    \multicolumn{1}{c}{} & 
    \multicolumn{1}{c}{Metal Piece} & 
    \multicolumn{1}{c}{Battery} &
    \multicolumn{1}{c}{PCB} &
    \multicolumn{1}{c}{Glass} \\ \hline
    \multirow{4}{*}{\rotatebox{90}{Actual\,}}
    & \multicolumn{1}{c|}{Metal Piece}  & 0 & 7 & 3 & 11\\ 
    & \multicolumn{1}{c|}{Battery}  &1   & 14 &6 &9 \\
    & \multicolumn{1}{c|}{PCB}  &3  & 3 & 7 &11 \\
    & \multicolumn{1}{c|}{Glass}  &1   & 1 & 3 &31 \\ \hline
    \end{tabular}
    \label{tab:matconf_fromscratch}
\end{table}

\begin{table}[t]
\renewcommand{\arraystretch}{1.3}
    \caption{Precision and recall for inference in each class for the randomly initialized model}
    \label{precision_recall_fromscratch}
    \begin{center}
    \begin{tabular}{c|c|c}
        \hline
        Class & Precision & Recall\\
        \hline
        Metal Piece & 0.00\% & 0.00\%\\ 
        Battery & 56.00\% & 46.67\% \\
        PCB  & 36.84\% & 29.17\%\\ 
        Glass & 50\% & 86.11\%\\
        \hline
    \end{tabular}
    \end{center}

\end{table}

Thus, the results with the randomly initialized network are drastically worse than the pre-trained network, justifying the choice to use a pre-trained model.

\subsubsection{Binary classification}
\begin{figure}[!t]
\centering
\includegraphics[width=3.2in]{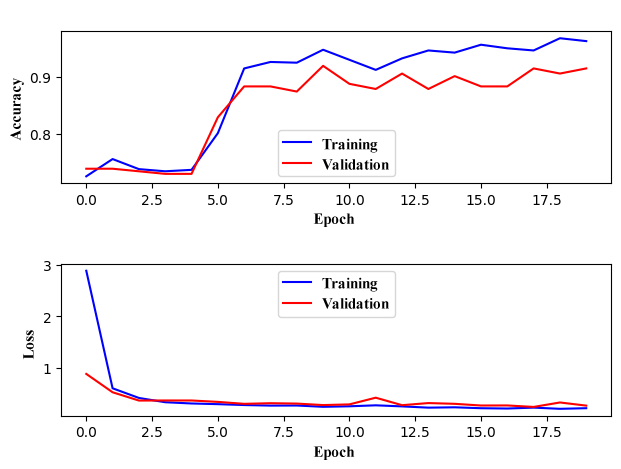}
\caption{Accuracy graphs (upper) and losses (bottom) of training and validation along the training.}
\label{fig:grafBinario}
\end{figure}

Because the work mainly emphasizes separating batteries from other materials, we also questioned if we could not optimize learning for only detecting what is and isn't a battery, especially considering that the battery class has the most visually consistent appearance. For that reason, we trained a version of the model with only two classes: Battery and Other (including PCBs, glasses, and metal pieces). This model was also pre-trained on ImageNet, and all training conditions except for the classes were equal to those of the first model.
The resulting model trained for a total 20 epochs, with the best one being epoch 10. It had a training accuracy of 94.71\% and loss of 0.2481, with a validation accuracy of 91.89\% and loss of 0.2816. The confusion matrix is shown in Table~\ref{matriz_confusao_binaria}, and the class-wise precision and recall are shown in Table~\ref{precision_recall_binaria}. The training graphs are shown in Fig.~\ref{fig:grafBinario}.

The drawback of the binary approach is that it is not possible to understand which component is contaminating the flow of batteries and therefore improve the performance of the model based on this information. As discussed in Section \ref{trainingresults}, if the only class being incorrectly sorted alongside batteries is metal pieces, then there is no impairment to the recycling process. However, pieces of glass would have potential to hinder the recyclability of the batteries, and PCBs would mean the loss of valuable elements.

\begin{table}[t]
    \centering
    \caption{Confusion matrix of the binary model's test predictions}
    \label{matriz_confusao_binaria}
    \begin{tabular}{cccc}
    \cline{3-4}
    \multicolumn{1}{c}{} &\multicolumn{1}{c}{} &\multicolumn{2}{c}{Prediction} \\ 
    \cline{3-4}
    && \multicolumn{1}{c}{Battery} & 
    \multicolumn{1}{c}{Other} \\ \hline
    \multirow{2}{*}{\rotatebox{90}{Act.\,\,}} 
    & \multicolumn{1}{|c|}{Battery}  & 26 & 4\\[.035in] 
    & \multicolumn{1}{|c|}{Other}  &0   & 81  \\
\hline
    \end{tabular}
    \label{tab:matconf_binary}
\end{table}

\begin{table}[t]
\renewcommand{\arraystretch}{1.3}
    \caption{Precision and recall for inference in each class for the binary model}
    \label{precision_recall_binaria}
    \begin{center}
    \begin{tabular}{c|c|c}
        \hline
        Class & Precision & Recall\\
        \hline
        Battery & 100.00\% & 86.67\% \\
        Other  & 95.29\% & 100.00\%\\ 
        \hline
    \end{tabular}
    \end{center}

\end{table}

\section{Conclusion} \label{sec_conclusions}

In this work, we present the possibility of using an image classification neural network as a cheaper and more efficient alternative in separating materials for recycling WEEE. We created an image dataset with 300 pictures of assorted components from pyrolyzed smartphones and an adapted version of this dataset for image classification with 1,127 individual images of these components. We then trained an image classification model on this dataset to differentiate between metal pieces, battery, PCB, and glass components, achieving an overall accuracy of 82.49\%.

The experiment results show promise for using a neural network to separate WEEEs, especially considering the small dataset used. Future works could expand the goal of this study to a detection problem, creating a model capable of both classifying and locating multiple components at once in real-time, which could even further allow for a compact and efficient automated separation system. The relatively high accuracies obtained with a lightweight network could also feasibly allow for low-cost yet accurate embedded systems to be developed for this purpose. Future works could look into testing other models---both alternative architectures and possibly detection models---as well as investigating the impact of the unbalanced dataset and ways to mitigate it, such as with class-specific augmentations or by using weighted loss functions.

Even considering the study's limitations, the model used achieved its purpose (separation of batteries) with high accuracies, with significant potential to be implemented on recycling routes and increase their effectiveness in the valorization of the residues. This approach can be implemented in a recycling line coupled with a mechanical sorting system. A camera pointing perpendicularly on a conveyor belt could feed a detection system that informs the mechanical devices to separate the target components. It is important to highlight that most electronics usually contain the same components as smartphones (metal pieces, glass, PCBs, and batteries). Thus, this approach has the potential to be applied to other types of devices, such as laptops and tablets. In addition to the simple separation of batteries, more studies can be carried out to refine the separation of the other classes of materials with high accuracy. 

Finally, the study created a dataset that could serve as a basis for the future development of models for similar cases. The fact that the dataset was annotated with polygonal masks also allows the annotations to be exported in different formats for different purposes, such as fine-grained semantic segmentation. We also intend to add more images to the dataset and, ultimately, make it public.



\section*{Acknowledgments}
We acknowledge the financial support from the Fundação de Amparo à Pesquisa do Estado do Rio Grande do Sul (FAPERGS - Brazil), the Conselho Nacional de Desenvolvimento Científico e Tecnológico (CNPq - Brazil) under the project UFRGS 43545, and the Coordenação de Aperfeiçoamento de Pessoal de Nível Superior (CAPES - Brazil) under the project PROEX 88887.501183/2020-00.




\bibliographystyle{unsrt}  
\bibliography{example}

\begin{thebibliography}{10}

\bibitem{e-waste2020}
V.~Forti, C.P. Baldé, R.~Kuehr, et~al.
\newblock {\em The Global E-waste Monitor 2020: Quantities, flows, and the circular economy potential}.
\newblock Bonn/Geneva/Rotterdam: UNU/UNITAR/ITU/ISWA, 2020.

\bibitem{umicoree-scrap}
Umicore Precious~Metals Refining.
\newblock E-scrap recycling.
\newblock Cited 16 August 2022.

\bibitem{industrialscale}
Muammer Kaya.
\newblock {\em Industrial-Scale E-Waste/WPCB Recycling Lines}, pages 177--209.
\newblock Springer International Publishing, Cham, 2019.

\bibitem{whatdrives}
P.R. Dias, M.P. Cenci, A.M. Bernardes, and N.~Huda.
\newblock What drives weee recycling? a comparative study concerning legislation, collection and recycling.
\newblock {\em Waste Management \& Research}, pages 1527--1538, 2022.

\bibitem{wasteinbrazil}
Pablo Dias, João Palomero, Marcelo~Pilotto Cenci, Tatiana Scarazzato, and Andréa~Moura Bernardes.
\newblock Electronic waste in brazil: Generation, collection, recycling and the covid pandemic.
\newblock {\em Cleaner Waste Systems}, 3:100022, 2022.

\bibitem{ecofriendly}
Marcelo~Pilotto Cenci, Tatiana Scarazzato, Daniel~Dotto Munchen, Paula~Cristina Dartora, Hugo~Marcelo Veit, Andrea~Moura Bernardes, and Pablo~R. Dias.
\newblock Eco-friendly electronics—a comprehensive review.
\newblock {\em Advanced Materials Technologies}, 7(2):2001263, 2022.

\bibitem{recoverymetals}
Muammer Kaya.
\newblock Recovery of metals and nonmetals from electronic waste by physical and chemical recycling processes.
\newblock {\em Waste Management}, 57:64--90, 2016.
\newblock WEEE: Booming for Sustainable Recycling.

\bibitem{pyro}
Marcus Sommerfeld, Claudia Vonderstein, Christian Dertmann, Jakub Klimko, Dušan Oráč, Andrea Miškufová, Tomáš Havlík, and Bernd Friedrich.
\newblock A combined pyro- and hydrometallurgical approach to recycle pyrolyzed lithium-ion battery black mass part 1: Production of lithium concentrates in an electric arc furnace.
\newblock {\em Metals}, 10(8), 2020.

\bibitem{computervisionwaste}
W.~Lu and J.~Chen.
\newblock Computer vision for solid waste sorting: A critical review of academic research.
\newblock {\em Waste Management \& Research}, 2022.

\bibitem{classificationsystem}
Janusz Bobulski and Mariusz Kubanek.
\newblock Waste classification system using image processing and convolutional neural networks.
\newblock In Ignacio Rojas, Gonzalo Joya, and Andreu Catala, editors, {\em Advances in Computational Intelligence}, pages 350--361, Cham, 2019. Springer International Publishing.

\bibitem{transferlearning}
Q~Zhang, Q~Yang, X~Zhang, et~al.
\newblock Waste image classification based on transfer learning and convolutional neural network.
\newblock {\em Waste Management \& Research}, 2021.

\bibitem{realtime}
Seok {Woo Yang}, Hyun {Joon Park}, Jin {Sob Kim}, Wonhee Choi, Jihwan Park, and Sung {Won Han}.
\newblock Study on the real-time object detection approach for end-of-life battery-powered electronics in the waste of electrical and electronic equipment recycling process.
\newblock {\em Waste Management}, 166:78--85, 2023.

\bibitem{automaticsorting}
Y~Lu, B~Yang, Y~Gao, et~al.
\newblock An automatic sorting system for electronic components detached from waste printed circuit boards.
\newblock {\em Waste Management \& Research}, 2022.

\bibitem{densenet}
Gao Huang, Zhuang Liu, Laurens Van Der~Maaten, and Kilian~Q Weinberger.
\newblock Densely connected convolutional networks.
\newblock In {\em Proceedings of the IEEE conference on computer vision and pattern recognition}, pages 4700--4708, 2017.

\bibitem{yolov4}
Alexey Bochkovskiy, Chien{-}Yao Wang, and Hong{-}Yuan~Mark Liao.
\newblock Yolov4: Optimal speed and accuracy of object detection.
\newblock {\em CoRR}, abs/2004.10934, 2020.

\bibitem{yolov3}
Joseph Redmon and Ali Farhadi.
\newblock Yolov3: An incremental improvement.
\newblock {\em CoRR}, abs/1804.02767, 2018.

\bibitem{roboflow}
B.~Dwyer, J.~Nelson, and J.~Solawetz.
\newblock Roboflow (version 1.0), 2022.

\bibitem{yi2021oriented}
Jingru Yi, Pengxiang Wu, Bo~Liu, Qiaoying Huang, Hui Qu, and Dimitris Metaxas.
\newblock Oriented object detection in aerial images with box boundary-aware vectors.
\newblock In {\em Proceedings of the IEEE/CVF Winter Conference on Applications of Computer Vision}, pages 2150--2159, 2021.

\bibitem{vgg}
Karen Simonyan and Andrew Zisserman.
\newblock Very deep convolutional networks for large-scale image recognition.
\newblock In Yoshua Bengio and Yann LeCun, editors, {\em 3rd International Conference on Learning Representations, {ICLR} 2015, San Diego, CA, USA, May 7-9, 2015, Conference Track Proceedings}, 2015.

\bibitem{smalldataset}
Shuying Liu and Weihong Deng.
\newblock Very deep convolutional neural network based image classification using small training sample size.
\newblock In {\em 2015 3rd IAPR Asian Conference on Pattern Recognition (ACPR)}, pages 730--734, 2015.

\bibitem{imagenet}
Karen Simonyan and Andrew Zisserman.
\newblock Very deep convolutional networks for large-scale image recognition.
\newblock In Yoshua Bengio and Yann LeCun, editors, {\em 3rd International Conference on Learning Representations, {ICLR} 2015, San Diego, CA, USA, May 7-9, 2015, Conference Track Proceedings}, 2015.

\bibitem{metalcasing}
Dana Thompson, Charlotte Hyde, Jennifer~M. Hartley, Andrew~P. Abbott, Paul~A. Anderson, and Gavin~D.J. Harper.
\newblock To shred or not to shred: A comparative techno-economic assessment of lithium ion battery hydrometallurgical recycling retaining value and improving circularity in lib supply chains.
\newblock {\em Resources, Conservation and Recycling}, 175:105741, 2021.

\bibitem{glass}
Ruixue Wang and Zhenming Xu.
\newblock Recycling of non-metallic fractions from waste electrical and electronic equipment (weee): A review.
\newblock {\em Waste Management}, 34(8):1455--1469, 2014.

\end{thebibliography}

\end{document}